\pdfoutput=1

\documentclass[11pt]{article}

\usepackage[]{ACL2023}

\usepackage{times}
\usepackage{latexsym}

\usepackage[T1]{fontenc}

\usepackage[utf8]{inputenc}
\usepackage{microtype}

\usepackage{inconsolata}

\usepackage{booktabs}
\usepackage{multirow}
\usepackage{graphicx}
\title{Understanding Code Semantics: An Evaluation of Transformer Models in Summarization}

\author{Debanjan Mondal$^{*}$ \quad Abhilasha Lodha$^{*}$ \quad Ankita Sahoo$^{*}$ \quad Beena Kumari$^{*}$\\
    University of Massachusetts Amherst\\
    \texttt{\{debanjanmond,alodha,asahoo,beenakumari\}@umass.edu} \\}
\begin{document}
\maketitle
\def\thefootnote{*}\footnotetext{Equal contribution.}
\begin{abstract}
This paper delves into the intricacies of code summarization using advanced transformer-based language models. Through empirical studies, we evaluate the efficacy of code summarization by altering function and variable names to explore whether models truly understand code semantics or merely rely on textual cues. We have also introduced adversaries like dead code and commented code across three programming languages (Python, Javascript, and Java) to further scrutinize the model's understanding. Ultimately, our research aims to offer valuable insights into the inner workings of transformer-based LMs, enhancing their ability to understand code and contributing to more efficient software development practices and maintenance workflows.
\end{abstract}

\section{Introduction}

Code summarization is a task that involves generating coherent and semantically relevant summaries that effectively describe the intended function of the software. In the dynamic realm of software development and maintenance, an adept grasp of program functionalities is of paramount importance. In this context, the integration of natural language summaries derived from source code emerges as a potent instrument, streamlining developers' efforts and augmenting program comprehension.

While current state-of-the-art code summarization models are developed and evaluated on clean and curated datasets, the real-world coding environment is far from standardized. Developers often deviate from standard coding practices, leading to inconsistent naming conventions. Additionally, actual codebases often feature commented sections, serving as legacy code or reserved for future use cases. Our research aims to simulate these real-world scenarios and assess whether models truly comprehend the inherent code semantics, rather than merely relying on textual cues.

The prevailing approaches to code summarization typically employ an encoder-decoder framework, encompassing the conversion of code into a hidden space and its subsequent transformation into natural language. For instance, CodeT5 \citep{wang-etal-2021-codet5}, a unified pretrained encoder-decoder Transformer model, leverages the semantics encoded in identifiers. In this research, we investgate the effectiveness of these models by tweaking the function and variable names in the existing code summarization datasets. Furthermore, we introduce additional challenges, such as commented code and dead code, to elevate the complexity of data samples and scrutinize the models' summarization processes. Dead code refers to unreachable code segments, devoid of functional importance, which language interpreters (e.g., Python and Javascript) ignore. We seek to evaluate whether models effectively disregard such code segments. All our experiments are reproducible and we will release our code and data upon publication.


The driving motivation behind this research lies in enhancing code comprehension and reducing the efforts entailed in software development and maintenance. By unraveling how Language Models comprehend code, we aim to contribute insights that pave the way for more effective software development practices. Our study, through experimentation and analysis, strives to provide valuable directions for improving the capabilities of Language Models in understanding and summarizing code, despite the challenges posed by real-world coding scenarios.\footnote{Our code is publicly available at: \href{https://github.com/Demon702/robust_code_summary}{Github}}

\section{Related Work}
Automated code summarization is a useful tool for software developers and has been an active reserach field for quite some time. Recently large language models have shown significant improvements in natural language tasks. Inspired by this, several pretrained language models have been developed for the programming language tasks. The encoder-decoder models have been found to be more successful in Programming Language (PL) tasks, whereas fully decoder models perform significantly better in Natural Language (NL) domain.

Models like CodeBERT \citep{feng-etal-2020-codebert}, PLBART \citep{ahmad-etal-2021-unified}, GraphCodeBERT \citep{guo2021graphcodebert}, CodeT5 \citep{wang-etal-2021-codet5}, CoTexT \citep{phan2021cotext} have shown impressive performances in the CodeXGLUE \cite{lu2021codexglue} benchmark. Unlike natural language, it's necessary to capture the rich code semantics in the programming language. Most Programming Language Models (PLMs) in this domain are pretrained on a large corpus of NL-PL pair in several programming languages with a masked token prediction objective. To capture the code semantics, various models have used different approaches. For example, CodeT5 uses an additional masked identifier prediction and GraphCodeBERT incorporates the data flow extracted from the code. These PLMs have shown impressive results in downstream tasks like code summarization. \citet{ahmed2022learning} explored the code summarization in project-specific domain. \citet{sun2022extractiveandabstractive} used an extractive and abstractive framework from source code summarization.
\citet{10.1145/3510003.3510049} showed that multilingual training can amply performance for low resource languages in different downstream tasks including code summarization. \citep{chen2022transferability} provided further insights for low resource languages like Ruby.

As indicated by \citet{guo2021graphcodebert} in GraphCodeBERT, indicators play a key role in code summarization. However, it's more desirable that our model relies more on code semantics and syntax rather than method names and identifiers. Developers follow their own naming conventions which can affect the model performance. In a closely related work, \citet{sontakke2022code} have shown that Semantic Preserving Transformations like removing code comments, replacing function names and local variable names to generic names significantly affects the BLEU score of the models like PLBART. We want to extend this exploration to other models like CodeT5, CodeBERT etc. We also aim to increase the scope of semantic preserving transformations by including dead code and commented code to check the model's understanding of the code.

 \renewcommand{\thefootnote}{\arabic{footnote}}
\section{Dataset}
There are several different code summarization datasets available. But we prefered CodeXGLUE \cite{lu2021codexglue} over others for these reasons - 
\begin{itemize}
    \item CodeXGLUE has been meticulously de-duplicated, as demonstrated in \citep{Shi_2022}. This ensures that any duplication within the dataset does not artificially inflate performance metrics.
    \item CodeXGLUE offers a wide range of six languages. This allowed us to conduct experiments and compare results across different programming languages.
\end{itemize}

We can categorize the six languages available in CodeXGLUE into three groups based on the size of their combined datasets (including train, validation, and test sets). Please note that the following information pertains to the combined dataset size:
\begin{itemize}
    \item In the High Resource category, both Python and PHP have approximately 300,000 code-summary pairs each.
    
    \item In the Mid Resource category, Java and Go consist of around 180,000 code-summary pairs.

    \item In the Low Resource category, Javascript comprises 65,000 pairs, while Ruby only has 27,000 code-summary pairs.
\end{itemize}
To analyze the impact of data transformation across resource categories, we selected one language from each category. Therefore, for our experiments, we utilized the languages Python, Java, and Javascript. The detailed statistics about train, validation and test splits is presented in Table \ref{tab:dataset-stats}.

\subsection{Data Transformation}
\label{sec:data-transformation}
\begin{figure*}[hbt!]
    \begin{center}
    \includegraphics[scale=0.5]{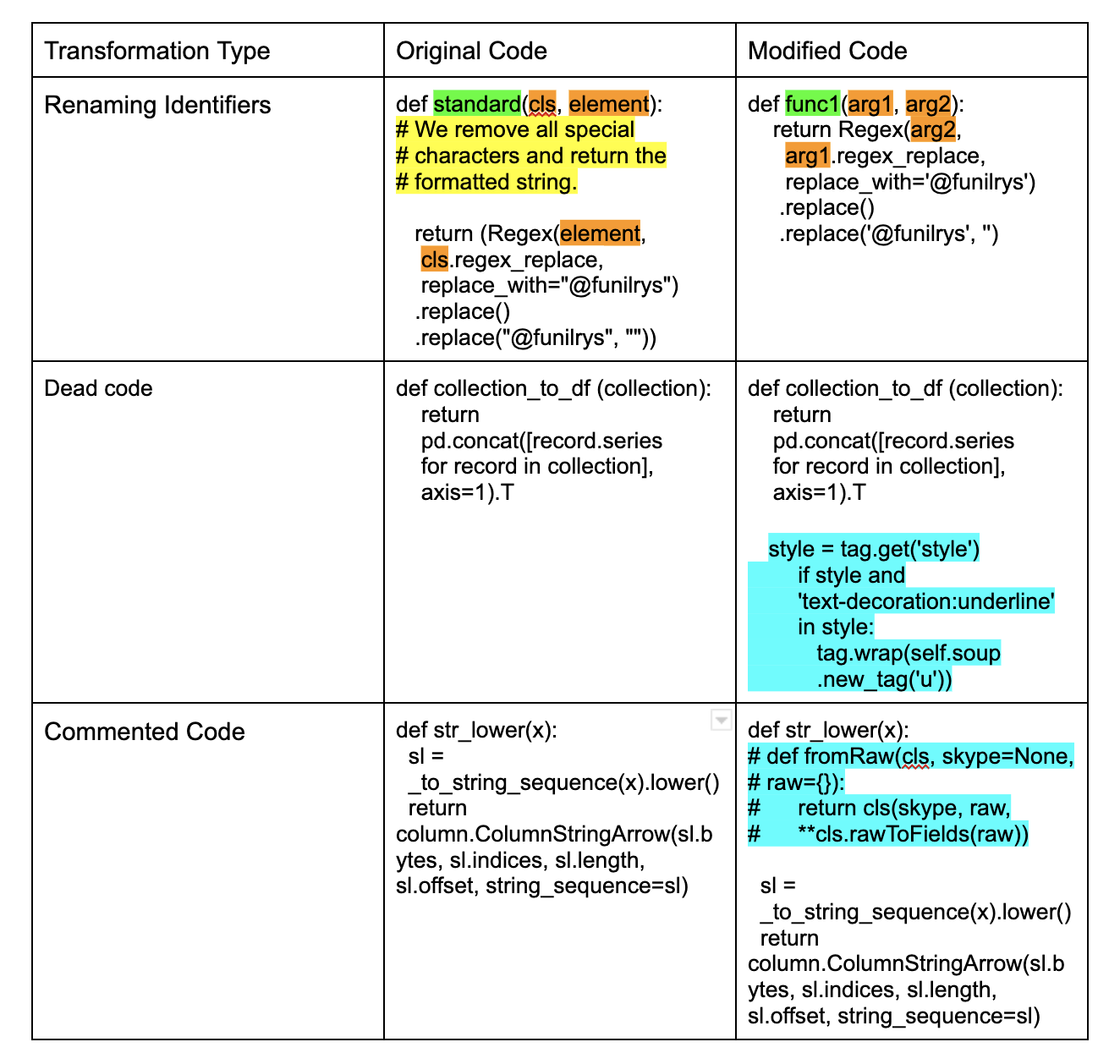}
    \caption{Examples of different kind of transformations discussed in Section \ref{sec:data-transformation}.}
    \label{fig:example}
    \end{center}
\end{figure*}

We will focus on code transformations that will preserve the code functionality. In the programming paradigm, this is known as \textbf{obfuscation}. In this study, we focused on 3 kinds of transformations. These are visually explained in Figure \ref{fig:example}:
\begin{itemize}
    \item \textbf{Renaming Identifiers:} Although the software development industry emphasizes the importance of meaningful and descriptive names for functions and variables, developers often use random function and variable names. To replicate such scenario, we replaced function and variable names with generic but unique names. However while doing so, we had to keep in mind that the control flow of the overall program should not be affected. We leveraged the Abstract Syntax Tree (AST) of the source code to identify and edit identifiers. Implementation details and interesting corner cases vary across programming languages and will be discussed subsequently.  
    \item \textbf{Commented Code:} It is very common in software engineering to encounter commented codes inside a function. These may be legacy codes which are not used anymore, or code snippets that might be used in future. To simulate such situation, we added commented codes. For each source code, we randomly sampled a function within the same data split, created commented version of it and added it after a function definition. Finding a suitable place to add comments is tricky and sometimes it can potentially change the program functionality. For our experiments, we add the comments starting from the next line after function definition.  
    \item \textbf{Dead Code:} Adding code after return statements is another transformation that we explored in our experiments. Python and Javascript interpreters ignore anything added after return statements, and we wanted to check if the models have developed the ability to do so. Since Java compiler throws error if we add anything after return, we excluded Java from this study. 
\end{itemize}

Code transformation implementation details about specific programming languages have been presented in \ref{appendix:data_transformation}.

\begin{table}[]
\centering
\resizebox{0.5\textwidth}{!}{%
\begin{tabular}{@{}llrrr@{}}
\toprule
\textbf{Languages}                   & \textbf{Transformation} & \textbf{Train Data} & \textbf{Validation Data} & \textbf{Test Data} \\ \midrule
\multirow{4}{*}{Python}     & Original               & 251820     & 13914           & 14918     \\
                            & Renamed Identifiers & 251820     & 13914           & 14918     \\
                            & Commented Code      & 251820     & 13914           & 14918     \\
                            & Dead Code           & 251820     & 13914           & 14918     \\ \midrule 
\multirow{4}{*}{Javascript} & Original               & 58025      & 3885            & 3291      \\
                            & Renamed Identifiers & 38254      & 2730            & 2157      \\
                            & Commented Code      & 38254      & 2730            & 2157      \\
                            & Dead Code           & 21897      & 1548            & 1213      \\  \midrule
Java                        & Original               & 164923     & 5183            & 10955     \\
                            & Renamed Identifiers & 164888     & 5182            & 10953     \\
                            & Commented Code      & 164923     & 5183            & 10955   \\ \bottomrule
\end{tabular}%
}
\caption{Dataset Size Information for different splits.}
\label{tab:dataset-stats}
\end{table}

\section{Models and Evaluation}

\subsection{Models}
In this study we ran our experiments on 2 models:
\begin{itemize}
    \item \textbf{CodeT5:} Upon examining the leaderboard of CodeXGLUE\footnote{\url{https://microsoft.github.io/CodeXGLUE/}}, and comparing metrics from various research papers, we discovered that CodeT5 \citep{wang-etal-2021-codet5} achieves state-of-the-art results for the code summarization task on this benchmark. Due to limitations in the size of our available GPUs, we opted to utilize the CodeT5 small\footnote{\url{https://huggingface.co/Salesforce/codet5-small}} (60M parameters) and base\footnote{\url{https://huggingface.co/Salesforce/codet5-base}} (223M parameters) models, while excluding the CodeT5 Large model (770M parameters) from our analysis. It is worth noting that CodeT5 incorporates an identifier aware denoising objective during its pretraining, making it more inclined to utilize textual cues from identifiers. We wanted to evaluate its robustness in our experiments.
    \item \textbf{CodeBERT:} For comparison across different architectures, we chose the CodeBERT \cite{feng-etal-2020-codebert} (173M parameters) model. Unlike CodeT5, CodeBERT is an encoder only model which is pretrained on the CodeSearchNet \citep{husain2020codesearchnet} dataset. For sequence-to-sequence generation problems like code summarization, the authors provide an Encoder-Decoder framework where they initialize the encoder with the pretrained CodeBERT, but they randomly initialize the decoder with a transformer model. Note that the decoder weights are not trained during the pretraining phase. For our experiments we use the CodeBERT Base\footnote{\url{https://huggingface.co/microsoft/codebert-base}} model.
\end{itemize}

The finetuning code utilized for our project was obtained from the public GitHub repository of CodeT5.\footnote{\url{https://github.com/salesforce/CodeT5}} The CodeT5 authors also included the finetuning code for CodeBERT. We reused the finetuned checkpoint of CodeT5 base that was available. As for CodeT5 small and CodeBERT, we performed the finetuning process on clean CodeXGLUE data, using the hyperparameters specified in the repository. The finetuned checkpoints on clean data for every model and every language serve as our baseline. Note that, the BLEU scores \citep{papineni-etal-2002-bleu} of our finetuned checkpoints on the clean CodeXGLUE test data has slight discrepancies with the BLEU scores reported in the original papers. However, the difference was not more than one decimal point. This may be attributed to the difference between GPU architectures. We report the scores mentioned in the original papers as the train clean - test clean scores.

\section{Experiments}
\label{sec:Experiments}
For all our experiments, we use the NVIDIA Tesla M40 GPUs. While finetuning for all types of data (Clean, Corrupted and Combined), we used the clean data finetuning hyperparameters that were available in the CodeT5 repository. However, for CodeT5 Base model, we had to reduce the batch size from 48 to 16 to fit in our GPU. All other hyperparameters remain the same. With this setup, we perform the following experiments.
\begin{itemize}
    \item \textbf{Identifier Corruption:} We conducted individual finetuning of the models using clean, corrupted, and combined train data (clean + corrupted). Subsequently, we evaluated these three types of models on both clean and corrupted test data. We used CodeBERT, CodeT5 Small and CodeT5 Base models for this experiment.

    \item \textbf{Commented Code Corruption:} For this corruption type, we performed separate finetuning of the model using clean data and commented code corrupted data. Finally, we evaluated these two types of models on both clean data and commented code corrupted data. We only evaluated the CodeT5 Small and CodeT5 base models in this particular setup.
    
    \item \textbf{Dead Code Corruption:} Similar to the previous corruption type, we carried out separate finetuning of the model using clean data and dead code corrupted data. Subsequently, we evaluated these two types of models on both clean data and dead code corrupted data. Once again, we only evaluated the CodeT5 Small and CodeT5 base models in this setup.
\end{itemize}

\subsection{Evaluation} 
Evaluation metrics are crucial for assessing the effectiveness of a code summarization model. In our study, we utilized both automatic and human evaluations.
\begin{enumerate}
    \item \textbf{BLEU} \cite{papineni-etal-2002-bleu}: It is a precision-based metric that measures the overlap between the words (and/or n-grams) in the machine-generated summaries and the human reference summaries. We calculated smoothed BLEU-4 (considering upto 4-grams) scores for each of the generated summaries and then averaged across all the summaries. We used the same evaluation script as CodeT5 which in turn reused the original evaluation script provided in the CodeXGLUE \cite{lu2021codexglue} benchmark. The BLEU scores obtained from different experiments for Python, Java, and JavaScript are presented in Table \ref{tab:identifier-corruption}, \ref{tab:dead-code}, and \ref{tab:commented-code}.
    \item \textbf{Human Evaluation:} Additionally, we performed a manual evaluation by annotating 200 random samples for both Python and JavaScript. Further details of the human evaluation process are discussed in the Section \ref{sec:Error Analysis}.
\end{enumerate}

\begin{table*}[!h]
\centering
\resizebox{\textwidth}{!}{%
\begin{tabular}{@{}llrrrrrrrrr@{}}
\toprule
\multirow{2}{*}{\textbf{Train Data}} &
  \multirow{2}{*}{\textbf{Test Data}} &
  \multicolumn{3}{c}{\textbf{Python}} &
  \multicolumn{3}{c}{\textbf{Javascript}} &
  \multicolumn{3}{c}{\textbf{Java}} \\ \cmidrule(l){3-5} \cmidrule(l){6-8} \cmidrule(l){9-11} 
 &
   &
  \textbf{CodeT5 Small} &
  \textbf{CodeBERT} &
  \textbf{CodeT5 Base} &
  \textbf{CodeT5 Small} &
  \textbf{CodeBERT} &
  \textbf{CodeT5 Base} &
  \textbf{CodeT5 Small} &
  \textbf{CodeBERT} &
  \textbf{CodeT5 Base} \\ \midrule
Clean     & Clean     & 19.96        & 19.06 & 20.01 & 15.32 & 14.90 & 16.16 & 20.02 & 17.65 & 20.31 \\
          & Corrupted & 12.92        & 13.03 & 12.81 & 9.51  & 7.34  & 9.04  & 15.06 & 13.42 & 14.05 \\ \midrule
Corrupted & Clean     & 19.28        & 17.75 & 19.64 & 14.55 & 11.30 & 15.41 & 19.05 & 17.16 & 19.50 \\
          & Corrupted & 16.21        & 15.52 & 16.50 & 12.68 & 10.76 & 13.13 & 17.27 & 16.20 & 17.36 \\ \midrule
Combined  & Clean     & 19.73        & 18.93 & 20.05 & 15.27 & 13.01 & 15.83 & 19.82 &    18.01   & 19.70 \\
          & Corrupted & 16.17 & 15.76 & 16.51 & 12.46 & 11.29 & 12.86 & 17.14 &   16.19    & 17.44 \\ \bottomrule
\end{tabular}%
}
\caption{Smooth BLEU-4 scores for different train-test combinations for clean and identifier corrupted data. When models are trained using clean data, their performance deteriorates when tested on identifier corrupted data. However, when models are trained on a combination of both clean and corrupted data, they demonstrate satisfactory performance on both types of test data - clean and corrupted.}
\label{tab:identifier-corruption}
\end{table*}

\begin{table}[!h]
\resizebox{0.5\textwidth}{!}{%
\begin{tabular}{@{}llrrrr@{}}
\toprule
\multirow{2}{*}{\textbf{Train Data}} & \multirow{2}{*}{\textbf{Test Data}} & \multicolumn{2}{c}{\textbf{Python}}          & \multicolumn{2}{c}{\textbf{Javascript}}      \\ \cmidrule(l){3-4} \cmidrule(l){5-6}
                                     &  & \textbf{CodeT5 Small} & \textbf{CodeT5 Base} & \textbf{CodeT5 Small} & \textbf{CodeT5 Base} \\ \midrule
Clean     & Clean     & 19.96 & 20.01 & 15.32 & 16.16 \\
          & Dead Code & 18.55 & 19.83 & 15.20 & 15.52 \\ \midrule
Dead Code & Clean     & 19.74 & 18.66 & 14.69 & 15.32 \\
          & Dead Code & 18.92 & 19.19 & 15.62 & 16.70 \\ \bottomrule
\end{tabular}%

}
\caption{Smooth BLEU-4 scores for different language and train-test combinations for dead code corruption.}
\label{tab:dead-code}
\end{table}

\begin{table*}[!h]
\centering
\resizebox{0.9\textwidth}{!}{%
\begin{tabular}{@{}llrrrrrr@{}}
\toprule
\multirow{2}{*}{\textbf{Train Data}} &
  \multirow{2}{*}{\textbf{Test Data}} &
  \multicolumn{2}{c}{\textbf{Python}} &
  \multicolumn{2}{c}{\textbf{Javascript}} &
  \multicolumn{2}{c}{\textbf{Java}} \\ \cmidrule(l){3-4} \cmidrule(l){5-6} \cmidrule(l){7-8} 
 &
   &
  \textbf{CodeT5 Small} &
  \textbf{CodeT5 Base} &
  \textbf{CodeT5 Small} &
  \textbf{CodeT5 Base} &
  \textbf{CodeT5 Small} &
  \textbf{CodeT5 Base} \\ \midrule
Clean     & Clean     & 19.96 & 20.01 & 15.32 & 16.16 & 20.02 & 20.31 \\
          & Commented & 16.15 & 16.26 & 14.61 & 14.21 & 15.06 & 18.83 \\
Commented & Clean     & 19.07 & 17.90 & 15.00 & 15.07 & 19.77 & 20.23 \\
          & Commented & 18.32 & 18.75 & 15.90 & 15.73 & 19.57 & 20.17 \\ \bottomrule
\end{tabular}%
}
\caption{Smooth BLEU-4 scores for different language and train-test combinations for commented code corruption. }
    \label{tab:commented-code}
\end{table*}
\section{Discussion}
Our experiments try to answer the following essential questions on models' understandability.
\subsection{Research Question 1: \textit{Does a model trained on clean data perform well on the identifier corrupted data?}} 
\begin{figure}[hbt!]
    \begin{center}
    \includegraphics[scale=0.5]{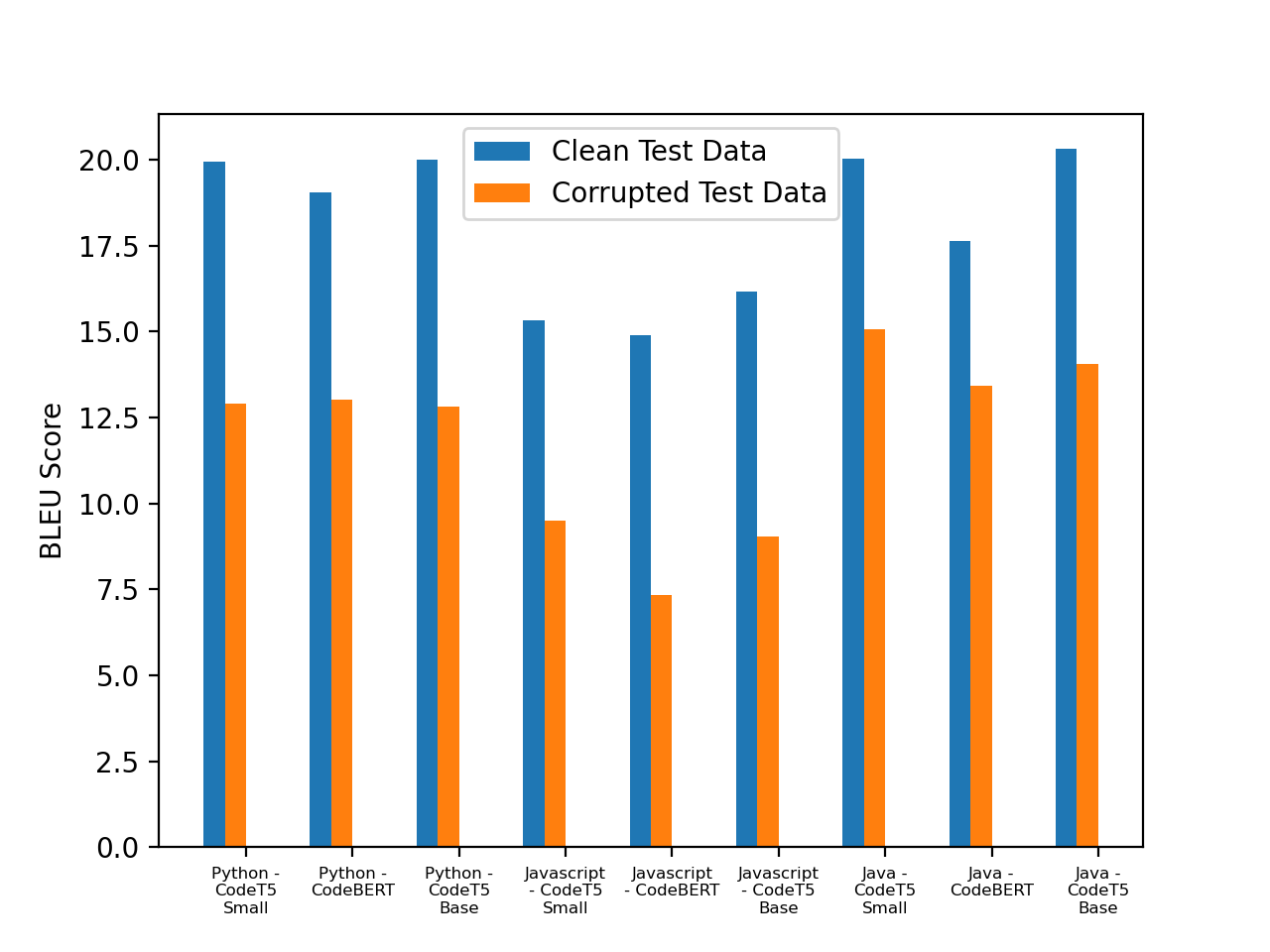}
    \caption{Comparing performance on clean and corrupted test data for models trained on clean data.}
    \label{fig:clean-vs-corrupted}
    \end{center}
\end{figure}

For all languages and models, the performance of a model trained on clean data tends to diminish when faced with corrupted test data, as compared to its performance on clean test data. The drop in performance for all models and languages is at least 4 points in terms of BLEU score. This phenomenon may be attributed to the model's reliance on textual hints found within function and variable names during training, rather than grasping the true essence of the code's functionality and achieving generalization. The comparison is visually shown in Figure \ref{fig:clean-vs-corrupted}.

\subsection{Research Question 2: \textit{Does the model trained on identifier corrupted data perform well on clean data?}}
\begin{figure}[hbt!]
    \begin{center}
    \includegraphics[scale=0.5]{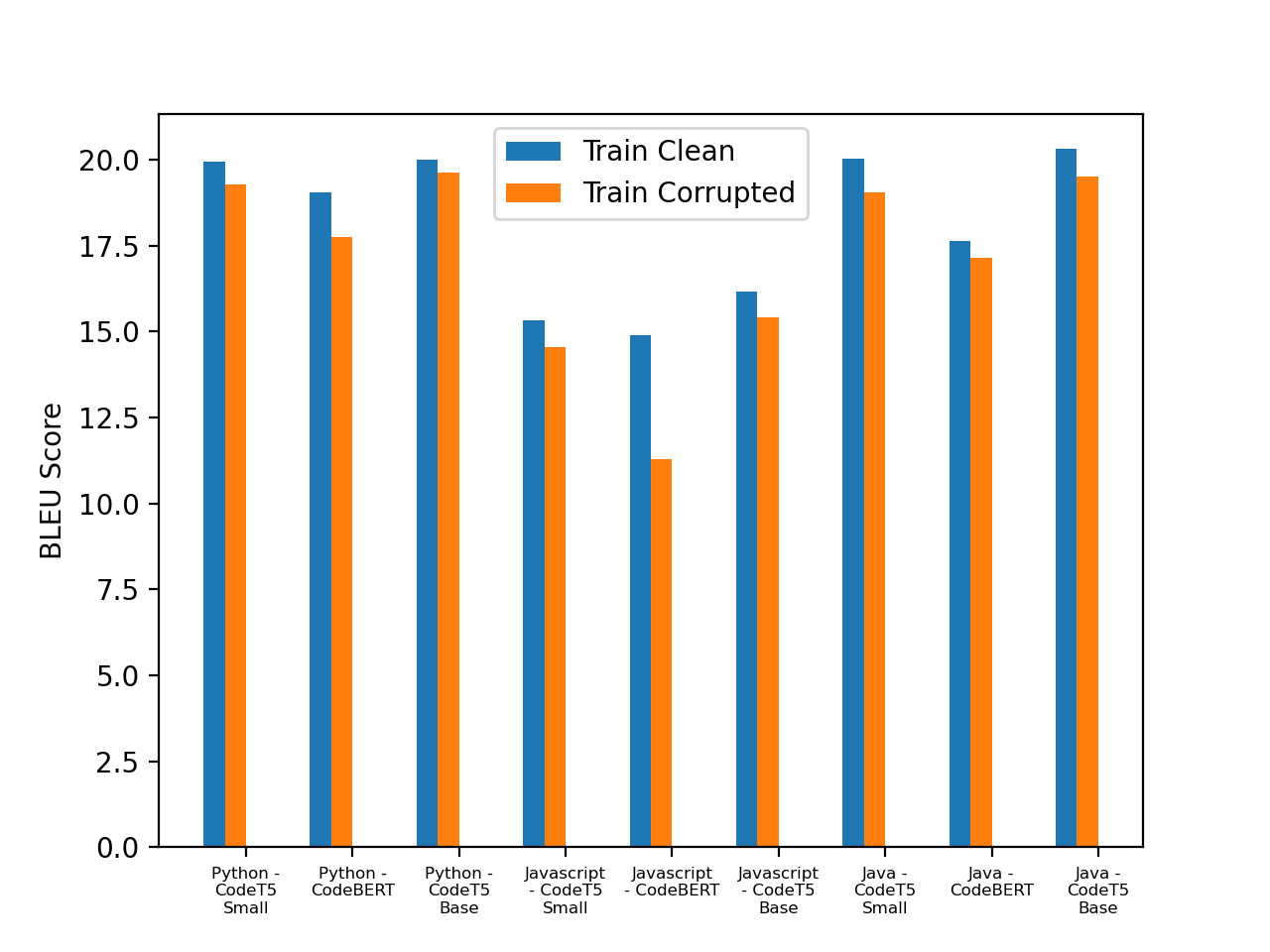}
    \caption{Comparing performance on clean test data and for models trained on clean data and corrupted data.}
    \label{fig:test-clean-train-clean-vs-corrupted}
    \end{center}
\end{figure}

A surprising observation arises when examining the performance of the model trained only on corrupted data. It demonstrates a commendable level of proficiency not only on corrupted test data (which is expected) but also on clean test data. This is visually explained in Figure \ref{fig:test-clean-train-clean-vs-corrupted}. For all languages and models, the performance on clean test data between the model trained on clean data and corrupted data is less than 1 point in terms of BLEU score, except for CodeBERT in javascript.
We hypothesize that when we train on the corrupted data, the model is forced to understand the code functionality in a generalized manner, thereby enabling it to perform well even in the clean dataset.

\subsection{Research Question 3: \textit{How is the performance of the model trained on the combined data?}}
\begin{figure}[hbt!]
    \begin{center}
    \includegraphics[scale=0.5]{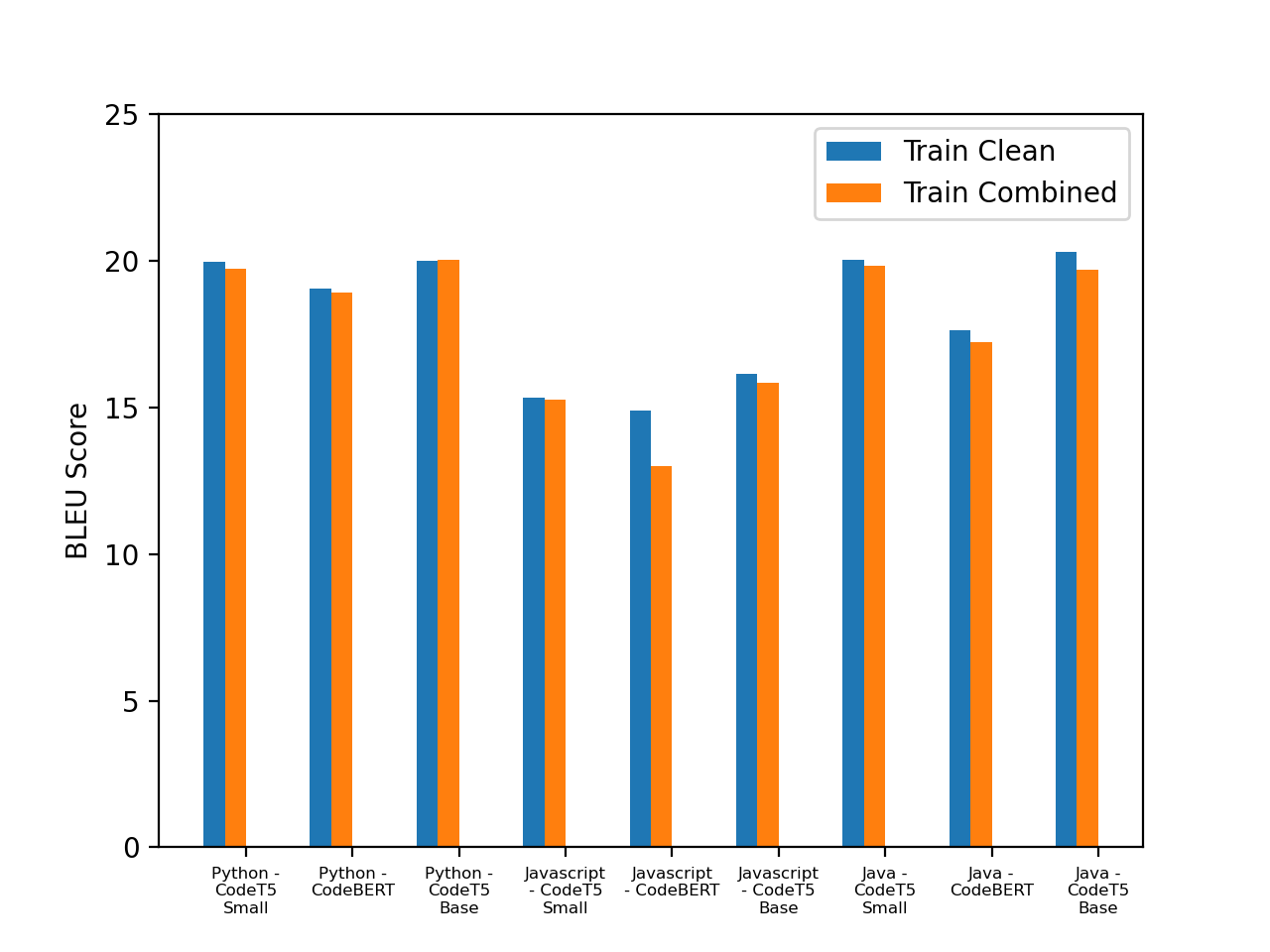}
    \caption{Comparing performance on clean test data for models trained on clean data and combined data.}
    \label{fig:test-clean-train-clean-vs-combined}
    \end{center}
\end{figure}

\begin{figure}[hbt!]
    \begin{center}
    \includegraphics[scale=0.5]{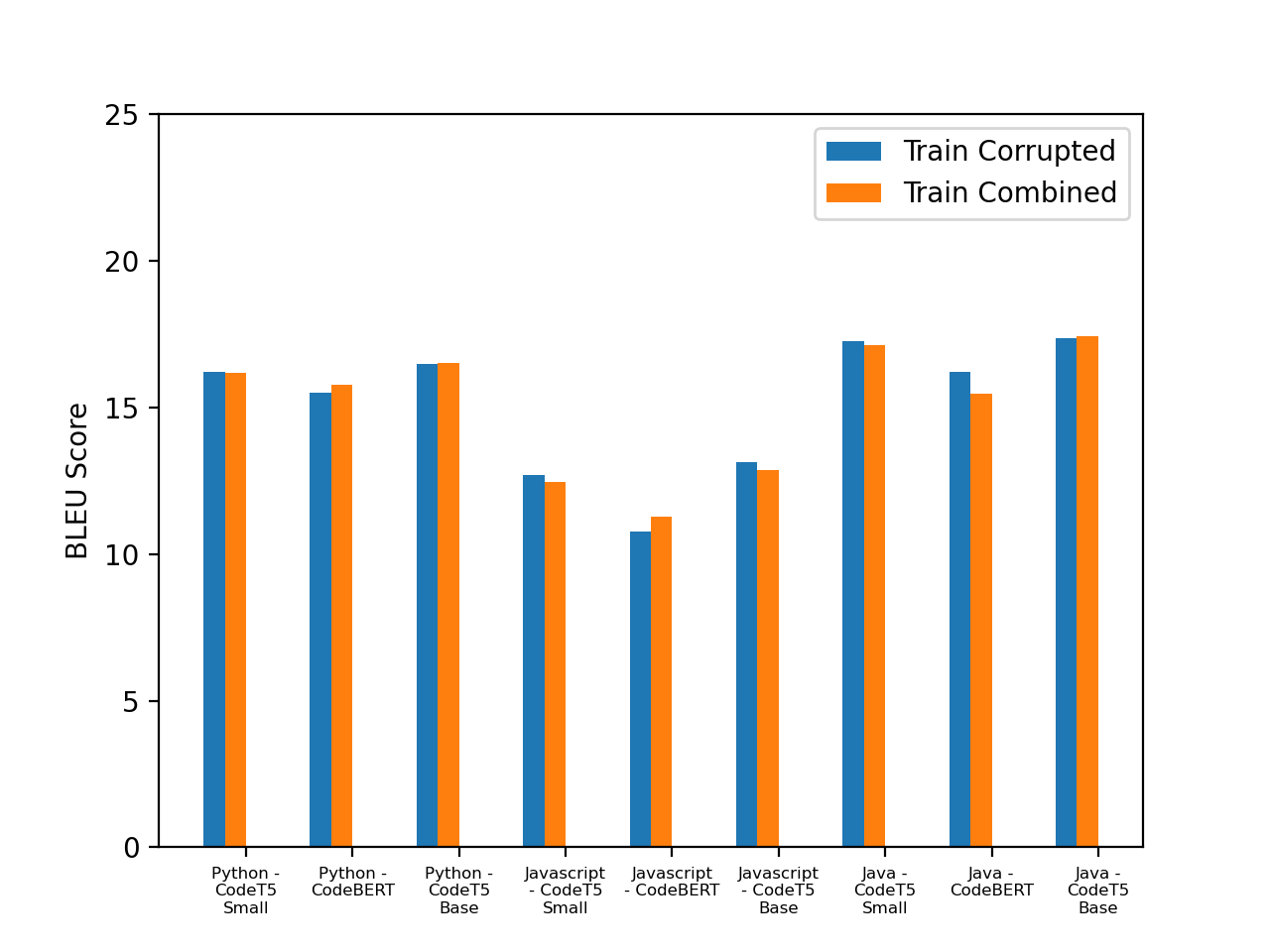}
    \caption{Comparing performance on corrupted test data for models trained on corrupted data and combined data.}
    \label{fig:test-corrupted-train-corrupted-vs-combined}
    \end{center}
\end{figure}

Our combined dataset contained both clean data and identifier corrupted data. The model trained on a combined dataset exhibits impressive performance not only on clean data but also on identifier corrupted data. Its performance on the clean test data is very similar to and sometimes even surpasses the performance of the model only trained on the clean data. Similar observations are seen for the corrupted test data. Notably, this pattern is consistent across all the models and languages. This shows that if we curate our dataset correctly, the model can generalize across clean and corrupted datasets. The BLEU score performance comparisons are visually explained in Figures \ref{fig:test-clean-train-clean-vs-combined} and \ref{fig:test-corrupted-train-corrupted-vs-combined}.

\subsection{Research Question 4: \textit{What is the effect of commented code perturbations on the model's capabilities?}}
\begin{figure}[hbt!]
    \begin{center}
    \includegraphics[scale=0.5]{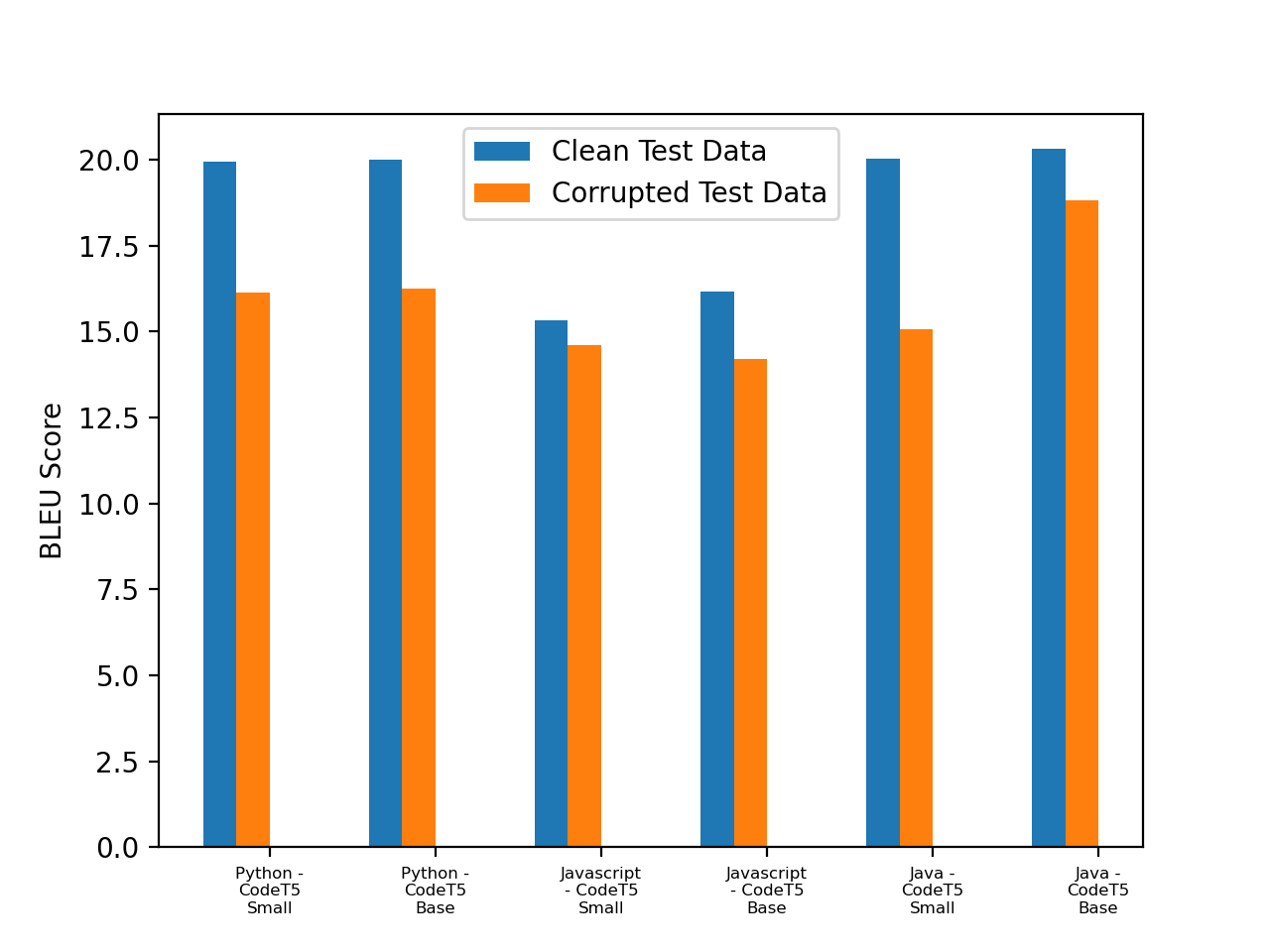}
    \caption{Comparing performance on clean and commented test data for models trained on clean data.}
    \label{fig:clean-vs-corrupted-commented}
    \end{center}
\end{figure}

We observe that the model trained on a dataset consisting of commented code showcases comparable and impressive performance on both clean code data and commented code data. However, a model trained exclusively on clean data displays satisfactory performance on the clean test set, albeit experiencing a notable decline in performance when evaluated on the commented code test set (presented in Figure \ref{fig:clean-vs-corrupted-commented}). A potential explanation is that the absence of comments in the clean code training data prevents the model from learning the syntax associated with comments. However, when the model is trained on code that includes comments, it captures the code syntax information despite the comments not directly influencing the code's functionality.

\subsection{Research Question 5: \textit{What is the effect of dead code perturbations on the model's capabilities?}}
Upon analysis, we determine that a model trained on a dataset that includes non-functional dead code demonstrates impressive and similar performance when applied to both clean code and the aforementioned dead code. We make the same observation for a model trained on clean data and evaluated on both clean and non-functional code test datasets. We thus conclude that the addition of dead code doesn't have any significant impact on the generated summaries.

\section{Human Evaluation}
\label{sec:Error Analysis}
In order to gain deeper insights into specific errors, a manual evaluation was conducted on a randomly selected subset of 200 examples from the identifier corrupted dataset. The evaluation compared the performance of two CodeT5 Base models: the model trained on the combined dataset (including identifier corrupted codes) and the baseline model trained on clean data. The evaluation involved analyzing code, gold truth, baseline model summaries, and combined data trained model summaries. We prepare one such set for both Python and Javascript (Refer Table \ref{tab:human-evaluation}). To avoid human bias, the annotators were given the summaries in a random order without any access to the information about which model generated them.

\begin{table}[!htbp]
\resizebox{0.5\textwidth}{!}{%
\begin{tabular}{@{}lrrrr@{}}
\toprule
\textbf{Language} &
  \multicolumn{1}{l}{\textbf{Clean Data Model}} &
  \multicolumn{1}{l}{\textbf{Combined Data Model}} &
  \multicolumn{1}{l}{\textbf{Ties}} &
  \multicolumn{1}{l}{\textbf{Total}} \\ \midrule
Python &
  21 &
  143 &
  36 &
  200 \\
Javascript &
  15 &
  148 &
  37 &
  200 \\ \bottomrule
\end{tabular}%
}
\caption{Statistics of the aggregated manual evaluation data determining which model's summary was better.}
    \label{tab:human-evaluation}
\end{table}

The evaluation process included two individuals independently annotating the same set of data and marking the annotations as \emph{Prediction 1}, \emph{Prediction 2}, or \emph{Tie} (both models). The chosen option implies that the selected annotation is more closely aligned with the gold truth and code. The inter-annotator agreement can be observed in the Table \ref{tab:agreement}. 

In cases where there was a disagreement between the annotators, \emph{Ties} were resolved using the following strategies:
\begin{itemize}
    \item If one annotation was marked as a \emph{Tie} and the other was marked as \emph{Prediction 1} or \emph{Prediction 2}, we considered the \emph{Prediction 1} or \emph{Prediction 2} annotation as the final annotation in the aggregated dataset.
    \item If one annotation was marked as \emph{Prediction 1} and the other as \emph{Prediction 2}, the two annotators engaged in a discussion to reach a consensus for the final annotation for the summaries.
\end{itemize}

The observations revealed that the summaries generated by the model trained on the combined data were more relevant to the code and closer to the desired outcome (gold truth) compared to the baseline model. There were several notable issues with the summaries generated by the baseline model, which are discussed in the Figure \ref{fig:example2}. 

The computed BLEU scores for the two models, where one was trained on clean data and the other on combined data, are as follows: For Python, the bleu scores on the identifier corrupted dataset are \textbf{12.92} and \textbf{16.17}, while for Javascript, the scores are \textbf{9.04} and \textbf{12.86}, respectively. These scores align with the manual evaluation results, indicating that the model trained on combined data outperforms the model trained solely on clean data in code summarization.
\section{Conclusion}
By studying advanced code summarization models, we discover how making changes that preserve the meaning of the code affects the quality of the summaries they generate. Additionally, we provide evidence that if we train the large language models like CodeT5 properly, making changes that disrupt the meaning of function and variable names have little impact on the resulting summaries. Our observations remain consistent across three distinct programming languages: Java, Python, and JavaScript. These findings raise important concerns about how well these models truly understand code, highlighting the need for better training methods and carefully curated datasets that improve their understanding. We propose using different types of code transformations, such as introducing renamed identifiers, adding comments, or dead code, as ways to enhance the training of these models. Furthermore, there is an exciting opportunity to apply these findings to different programming languages, so that we can learn more about their general applicability.

\section*{Limitations}
While this study presents valuable insights into code summarization using CodeBERT and CodeT5, certain limitations merit consideration.

Firstly, the experimentation focused exclusively on CodeBERT and CodeT5 due to practical GPU restrictions. While Large Language Model (LLMs) based approaches hold immense potential, their exclusion from the evaluation due to GPU limitations might restrict the generalizability of findings.

Secondly, the reliance on BLEU evaluation metrics, although widely used, introduces its own limitations\citep{10.1145/3468264.3468588}. BLEU captures the word-level overlap between generated and reference summaries, but it may not holistically reflect the quality of the summary in all cases. The intricate semantics and contextual intricacies present in code may not be fully captured by BLEU scores alone.

Moreover, while human evaluation was conducted on a subset of 200 samples, the comprehensiveness of this assessment could have been further extended. A more expansive human evaluation, covering a broader array of code samples, could provide a richer understanding of the models' actual performance.

In future studies, overcoming these limitations could involve wider experimentation across a spectrum of Language Models, a more robust human evaluation, and exploring alternative evaluation metrics that better align with the complex nature of code summarization.

\bibliography{anthology,custom}

\begin{thebibliography}{15}
\expandafter\ifx\csname natexlab\endcsname\relax\def\natexlab#1{#1}\fi

\bibitem[{Ahmad et~al.(2021)Ahmad, Chakraborty, Ray, and
  Chang}]{ahmad-etal-2021-unified}
Wasi Ahmad, Saikat Chakraborty, Baishakhi Ray, and Kai-Wei Chang. 2021.
\newblock \href {https://doi.org/10.18653/v1/2021.naacl-main.211} {Unified
  pre-training for program understanding and generation}.
\newblock In \emph{Proceedings of the 2021 Conference of the North American
  Chapter of the Association for Computational Linguistics: Human Language
  Technologies}, pages 2655--2668, Online. Association for Computational
  Linguistics.

\bibitem[{Ahmed and Devanbu(2022{\natexlab{a}})}]{ahmed2022learning}
Toufique Ahmed and Premkumar Devanbu. 2022{\natexlab{a}}.
\newblock \href {http://arxiv.org/abs/2206.00804} {Learning code summarization
  from a small and local dataset}.

\bibitem[{Ahmed and Devanbu(2022{\natexlab{b}})}]{10.1145/3510003.3510049}
Toufique Ahmed and Premkumar Devanbu. 2022{\natexlab{b}}.
\newblock \href {https://doi.org/10.1145/3510003.3510049} {Multilingual
  training for software engineering}.
\newblock In \emph{Proceedings of the 44th International Conference on Software
  Engineering}, ICSE '22, page 1443–1455, New York, NY, USA. Association for
  Computing Machinery.

\bibitem[{Chen et~al.(2022)Chen, Fard, Lo, and
  Bryksin}]{chen2022transferability}
Fuxiang Chen, Fatemeh Fard, David Lo, and Timofey Bryksin. 2022.
\newblock \href {http://arxiv.org/abs/2204.09653} {On the transferability of
  pre-trained language models for low-resource programming languages}.

\bibitem[{Feng et~al.(2020)Feng, Guo, Tang, Duan, Feng, Gong, Shou, Qin, Liu,
  Jiang, and Zhou}]{feng-etal-2020-codebert}
Zhangyin Feng, Daya Guo, Duyu Tang, Nan Duan, Xiaocheng Feng, Ming Gong, Linjun
  Shou, Bing Qin, Ting Liu, Daxin Jiang, and Ming Zhou. 2020.
\newblock \href {https://doi.org/10.18653/v1/2020.findings-emnlp.139}
  {{C}ode{BERT}: A pre-trained model for programming and natural languages}.
\newblock In \emph{Findings of the Association for Computational Linguistics:
  EMNLP 2020}, pages 1536--1547, Online. Association for Computational
  Linguistics.

\bibitem[{Guo et~al.(2021)Guo, Ren, Lu, Feng, Tang, LIU, Zhou, Duan,
  Svyatkovskiy, Fu, Tufano, Deng, Clement, Drain, Sundaresan, Yin, Jiang, and
  Zhou}]{guo2021graphcodebert}
Daya Guo, Shuo Ren, Shuai Lu, Zhangyin Feng, Duyu Tang, Shujie LIU, Long Zhou,
  Nan Duan, Alexey Svyatkovskiy, Shengyu Fu, Michele Tufano, Shao~Kun Deng,
  Colin Clement, Dawn Drain, Neel Sundaresan, Jian Yin, Daxin Jiang, and Ming
  Zhou. 2021.
\newblock \href {https://openreview.net/forum?id=jLoC4ez43PZ}
  {Graphcode{\{}bert{\}}: Pre-training code representations with data flow}.
\newblock In \emph{International Conference on Learning Representations}.

\bibitem[{Husain et~al.(2020)Husain, Wu, Gazit, Allamanis, and
  Brockschmidt}]{husain2020codesearchnet}
Hamel Husain, Ho-Hsiang Wu, Tiferet Gazit, Miltiadis Allamanis, and Marc
  Brockschmidt. 2020.
\newblock \href {http://arxiv.org/abs/1909.09436} {Codesearchnet challenge:
  Evaluating the state of semantic code search}.

\bibitem[{Lu et~al.(2021)Lu, Guo, Ren, Huang, Svyatkovskiy, Blanco, Clement,
  Drain, Jiang, Tang, Li, Zhou, Shou, Zhou, Tufano, GONG, Zhou, Duan,
  Sundaresan, Deng, Fu, and LIU}]{lu2021codexglue}
Shuai Lu, Daya Guo, Shuo Ren, Junjie Huang, Alexey Svyatkovskiy, Ambrosio
  Blanco, Colin Clement, Dawn Drain, Daxin Jiang, Duyu Tang, Ge~Li, Lidong
  Zhou, Linjun Shou, Long Zhou, Michele Tufano, MING GONG, Ming Zhou, Nan Duan,
  Neel Sundaresan, Shao~Kun Deng, Shengyu Fu, and Shujie LIU. 2021.
\newblock \href {https://openreview.net/forum?id=6lE4dQXaUcb} {Code{XGLUE}: A
  machine learning benchmark dataset for code understanding and generation}.
\newblock In \emph{Thirty-fifth Conference on Neural Information Processing
  Systems Datasets and Benchmarks Track (Round 1)}.

\bibitem[{Papineni et~al.(2002)Papineni, Roukos, Ward, and
  Zhu}]{papineni-etal-2002-bleu}
Kishore Papineni, Salim Roukos, Todd Ward, and Wei-Jing Zhu. 2002.
\newblock \href {https://doi.org/10.3115/1073083.1073135} {{B}leu: a method for
  automatic evaluation of machine translation}.
\newblock In \emph{Proceedings of the 40th Annual Meeting of the Association
  for Computational Linguistics}, pages 311--318, Philadelphia, Pennsylvania,
  USA. Association for Computational Linguistics.

\bibitem[{Phan et~al.(2021)Phan, Tran, Le, Nguyen, Anibal, Peltekian, and
  Ye}]{phan2021cotext}
Long Phan, Hieu Tran, Daniel Le, Hieu Nguyen, James Anibal, Alec Peltekian, and
  Yanfang Ye. 2021.
\newblock \href {http://arxiv.org/abs/2105.08645} {Cotext: Multi-task learning
  with code-text transformer}.

\bibitem[{Roy et~al.(2021)Roy, Fakhoury, and
  Arnaoudova}]{10.1145/3468264.3468588}
Devjeet Roy, Sarah Fakhoury, and Venera Arnaoudova. 2021.
\newblock \href {https://doi.org/10.1145/3468264.3468588} {Reassessing
  automatic evaluation metrics for code summarization tasks}.

\bibitem[{Shi et~al.(2022)Shi, Wang, Du, Chen, Han, Zhang, Zhang, and
  Sun}]{Shi_2022}
Ensheng Shi, Yanlin Wang, Lun Du, Junjie Chen, Shi Han, Hongyu Zhang, Dongmei
  Zhang, and Hongbin Sun. 2022.
\newblock \href {https://doi.org/10.1145/3510003.3510060} {On the evaluation of
  neural code summarization}.
\newblock In \emph{Proceedings of the 44th International Conference on Software
  Engineering}. {ACM}.

\bibitem[{Sontakke et~al.(2022)Sontakke, Patwardhan, Vig, Medicherla, Naik, and
  Shroff}]{sontakke2022code}
Ankita~Nandkishor Sontakke, Manasi Patwardhan, Lovekesh Vig, Raveendra~Kumar
  Medicherla, Ravindra Naik, and Gautam Shroff. 2022.
\newblock \href {https://openreview.net/forum?id=rI5ll2_-1Zc} {Code
  summarization: Do transformers really understand code?}
\newblock In \emph{Deep Learning for Code Workshop}.

\bibitem[{Sun et~al.(2022)Sun, Fang, Chen, Zhang, Tao, Han, Ge, You, and
  Luo}]{sun2022extractiveandabstractive}
Weisong Sun, Chunrong Fang, Yuchen Chen, Quanjun Zhang, Guanhong Tao, Tingxu
  Han, Yifei Ge, Yudu You, and Bin Luo. 2022.
\newblock \href {http://arxiv.org/abs/2206.07245} {An
  extractive-and-abstractive framework for source code summarization}.

\bibitem[{Wang et~al.(2021)Wang, Wang, Joty, and Hoi}]{wang-etal-2021-codet5}
Yue Wang, Weishi Wang, Shafiq Joty, and Steven~C.H. Hoi. 2021.
\newblock \href {https://doi.org/10.18653/v1/2021.emnlp-main.685} {{C}ode{T}5:
  Identifier-aware unified pre-trained encoder-decoder models for code
  understanding and generation}.
\newblock In \emph{Proceedings of the 2021 Conference on Empirical Methods in
  Natural Language Processing}, pages 8696--8708, Online and Punta Cana,
  Dominican Republic. Association for Computational Linguistics.

\end{thebibliography}
\bibliographystyle{acl_natbib}

\appendix

\newpage
\section{Appendix}

\subsection{Additional information on human annotation}
\begin{table}[!htbp]
\centering
\resizebox{0.5\textwidth}{!}{%
\begin{tabular}{@{}lrr@{}}
\toprule
\textbf{Language} & \multicolumn{1}{l}{\textbf{Raw Agreement}} & \multicolumn{1}{l}{\textbf{Cohen's Kappa}} \\ \midrule
Python            & 74.5\%                                     & 0.62                                       \\
Javascript        & 82.0\%                                     & 0.73                                       \\ \bottomrule
\end{tabular}%
}
\caption{Inter-annotator Agreement Statistics.}
\label{tab:agreement}
\end{table}


\begin{figure*}[hbt!]
    \begin{center}
    \includegraphics[scale=0.5]{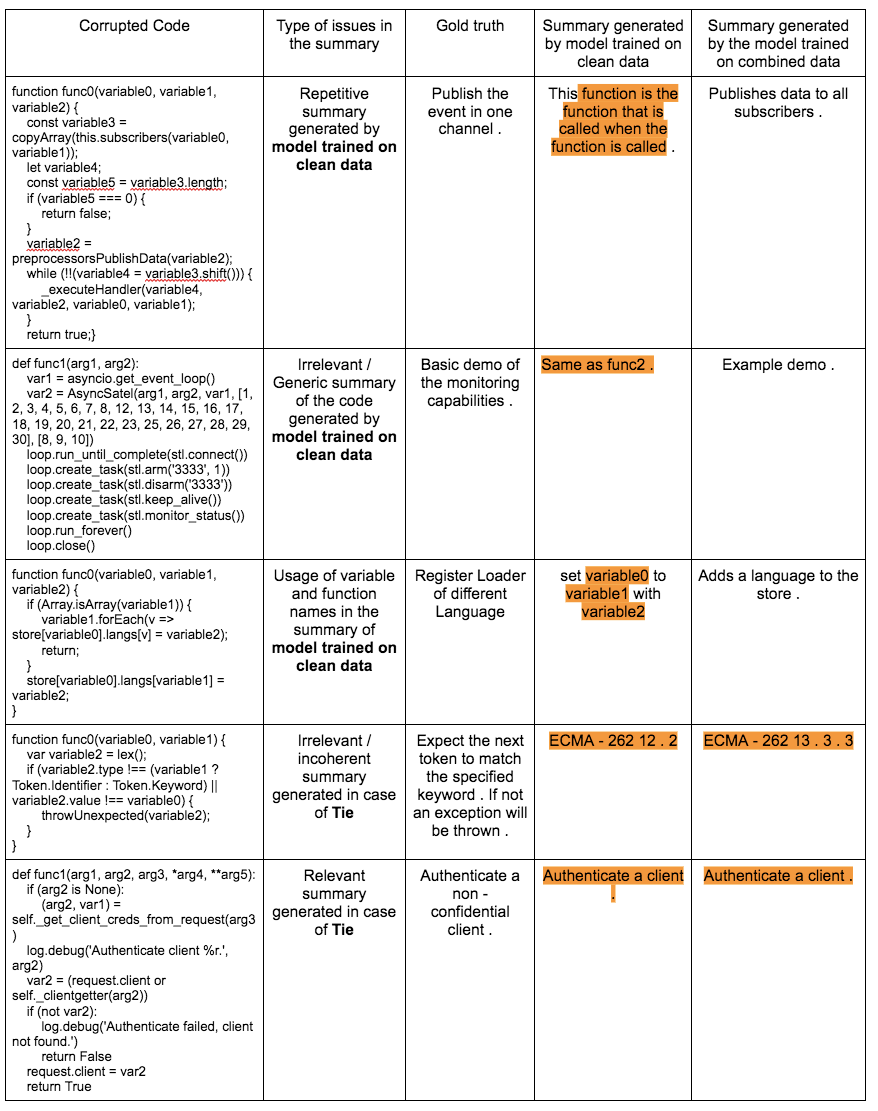}
    \caption{Examples of different summaries generated by the models in manual evaluation.}
    \label{fig:example2}
    \end{center}
\end{figure*}

\subsection{Data Transformation Procedure}
\label{appendix:data_transformation}
\subsubsection{Python}
\begin{itemize}
    \item \textbf{Renaming Identifiers:} We use Python's \emph{ast} package to parse the code and create AST. Then we transform the ast using \emph{NodeTransformer}\footnote{\url{https://docs.python.org/3/library/ast.html\#ast.NodeTransformer}} class. Finally the modified ast is unparsed and saved to the output file.
    \item \textbf{Commented Code:} Commented code is added after the function definition by randomly selecting a function code snippet from the same data split and adding comment symbols (\#) before each line of the selected code snippet.
    \item \textbf{Dead Code:} The code in this section adds extra code snippets after return statements in Python source code. The extra code snippet is taken from the function body of a randomly selected function in the same data split. It uses the \emph{libcst}\footnote{\url{https://libcst.readthedocs.io/en/latest/}} library to identify the location of the return statement.
\end{itemize}

\subsubsection{Javascript}
\begin{itemize}
    \item \textbf{Renaming Identifiers:} To achieve this, the \emph{esprima}\footnote{\url {https://www.npmjs.com/package/esprima}} library was employed to obtain the AST structure. Some codes are excluded from our dataset because the library fails to obtain the AST for those specific codes. The AST was traversed using \emph{Depth-First Search (DFS)} to extract node details of Identifiers related to variables and functions. Subsequently, the \emph{estraverse}\footnote{\url{https://www.npmjs.com/package/@types/estraverse}} library was utilized to traverse the AST and rename each identified node accordingly. Finally, the modified code was generated using \emph{escodegen}\footnote{\url{https://www.npmjs.com/package/escodegen}} and saved as the output. 
    \item \textbf{Commented Code:} The commented code is inserted following the function signature. This is performed after getting the AST and using the AST to identify the end of function signature. The code used for commenting is chosen randomly from a collection of code snippets found in another JSON object, with comment symbols (//) added to each line.
    \item \textbf{Dead Code:} Similar to commented code example, a function body code is randomly selected from a collection of code snippets in the same data split. This code is then appended to the original code after the return statement. 
\end{itemize}
\subsubsection{Java}
\begin{itemize}
    \item \textbf{Renaming Identifiers:} To modify the function names, and variable names, AST was generated for each code input samples using \emph{JavaParser}\footnote{\url{https://javaparser.org/}} package. The AST was traversed to extract the function and variable name nodes, which was then modified to generalized names. It is taken care of to replace the occurrence of same variable and function names with the modified name throughout the code sample using a hash map.
    \item \textbf{Commented Code:}  For adding commented code, we searched for the first opening curly braces "\{" and the commented code was inserted within /* ... */ and added after the aforementioned curly braces. The commented code snippets were randomly sampled from the same data split of Java code samples.
    \item \textbf{Dead Code:} Addition of codes after return statement in Java throws compile error, therefore dead code was not added to Java code samples.
    
\end{itemize}

\end{document}